\documentclass[twoside,11pt]{article}

\usepackage{automl2020}

\usepackage{booktabs} 

\usepackage[utf8]{inputenc}
\usepackage[T1]{fontenc}
\usepackage{alltt}
\usepackage{enumitem}
\usepackage[para]{footmisc}
\usepackage{graphicx}
\usepackage{listings}
\usepackage{url}
\usepackage{booktabs}
\usepackage{xcolor}
\usepackage{cleveref}
\usepackage{wrapfig}
\usepackage{zi4}
\usepackage{xspace}

\setlist{
  topsep=0mm,
  itemsep=0mm,
}

\lstdefinelanguage{bnf}{
  xleftmargin=3mm,
  basicstyle=\footnotesize,
  identifierstyle=\itshape,
  stringstyle=\ttfamily,
  columns=flexible,
  numbers=none,
  moredelim=[is][\footnotesize\sffamily]{'}{'},
  moredelim=[is][\footnotesize\sffamily]{"}{"},
}
\newcommand{\bnf}[1]{\lstinline[language=bnf]{#1}}

\lstdefinelanguage{python}{
  morestring=[b]',
  morestring=[b]""",
  morestring=[b]",
  morecomment=[l]\#,
  stringstyle=\ttfamily\color{black!60!white},
  showstringspaces=false,
  morekeywords={self,and,as,assert,break,class,continue,def,del,elif,else,except,False,finally,for,from,global,if,import,in,is,lambda,None,nonlocal,not,or,pass,raise,return,True,try,while,with,yield},
  keywordstyle=\color{blue!50!black}, 
}
\newcommand{\python}[1]{\lstinline[language=python]{#1}}

\lstdefinelanguage{desc}{
  basicstyle=\small\sffamily\color{black!60!white}
}

\lstdefinelanguage{json}{
  morekeywords={additionalProperties,allOf,anyOf,default,description,distribution,enum,exclusiveMaximum,exclusiveMinimum,maximum,maximumForOptimizer,minimum,minimumForOptimizer,not,properties,relevantToOptimizer,required,type},
  keywordstyle=\color{blue!50!black}
}
\newcommand{\json}[1]{\lstinline[language=json]{#1}}

\renewcommand{\cite}[1]{\citep{#1}}

\newcommand{\openmlDatasets}{15\xspace}
\newcommand{\totalOperators}{119\xspace}
\newcommand{\sklearnOperators}{115\xspace}
\newcommand{\curatedOperators}{42\xspace}
\newcommand{\sklearnCuratedOperators}{38\xspace}

\title{Mining Documentation to Extract Hyperparameter Schemas}

\author{\name Guillaume Baudart \and
        \name{Peter D.\ Kirchner} \and 
        \name{Martin Hirzel} \and
        \name{Kiran Kate}
       \addr IBM Research, New York, USA}
       
\jmlrheading{G. Baudart, P. Kirchner, M. Hirzel, K. Kate}
\ShortHeadings{Mining Documentation to Extract Hyperparameter Schemas}{G. Baudart, P. Kirchner, M. Hirzel, K. Kate}

\firstpageno{1}

\begin{document}
\maketitle

\begin{abstract}
AI automation tools need machine-readable hyperparameter schemas to
define their search spaces. At the same time, AI libraries often come
with good human-readable documentation. While such documentation
contains most of the necessary information, it is unfortunately not ready to
consume by tools.  This paper describes how to automatically mine
Python docstrings in AI libraries to extract JSON Schemas for their
hyperparameters. We evaluate our approach on \totalOperators transformers and
estimators from three different libraries and find that it is
effective at extracting machine-readable schemas.
Our vision is to reduce the burden to manually create
and maintain such schemas for AI automation tools and broaden the
reach of automation to larger libraries and richer schemas.

\end{abstract}

\section{Introduction}\label{sec:intro}

Machine-learning practitioners use libraries of \emph{operators}:
reusable implementations of estimators (such as logistic regression,
LR) and transformers (such as principal component analysis, PCA).
Training an operator fits its \emph{parameters} (learnable
coefficients such as LR weights or PCA eigenvectors) to a dataset.
Besides parameters, most operators also have \emph{hyperparameters}:
arguments that must be configured before training, such as the choice
of LR solver or the number of PCA components.  Python libraries for
machine learning (ML) such as scikit-learn~\cite{buitinck_et_al_2013} tend
to have good human readable documentation for hyperparameters.
Unfortunately, this documentation is usually not easily machine
readable.
ML practitioners can configure hyperparameters either by hand or by
using an HPO (automated hyperparameter optimization) tool such as
hyperopt-sklearn~\cite{komer_bergstra_eliasmith_2014} or
auto-sklearn~\cite{feurer_et_al_2015},
or the grid search or randomized search from
scikit-learn.  A \emph{hyperparameter schema}
specifies which hyperparameters are categorical and which continuous,
which values or ranges are valid, and conditional hyperparameter
constraints.  


\medskip

\begin{wrapfigure}{r}{0.55\textwidth}
  \centerline{\includegraphics[scale=0.45]{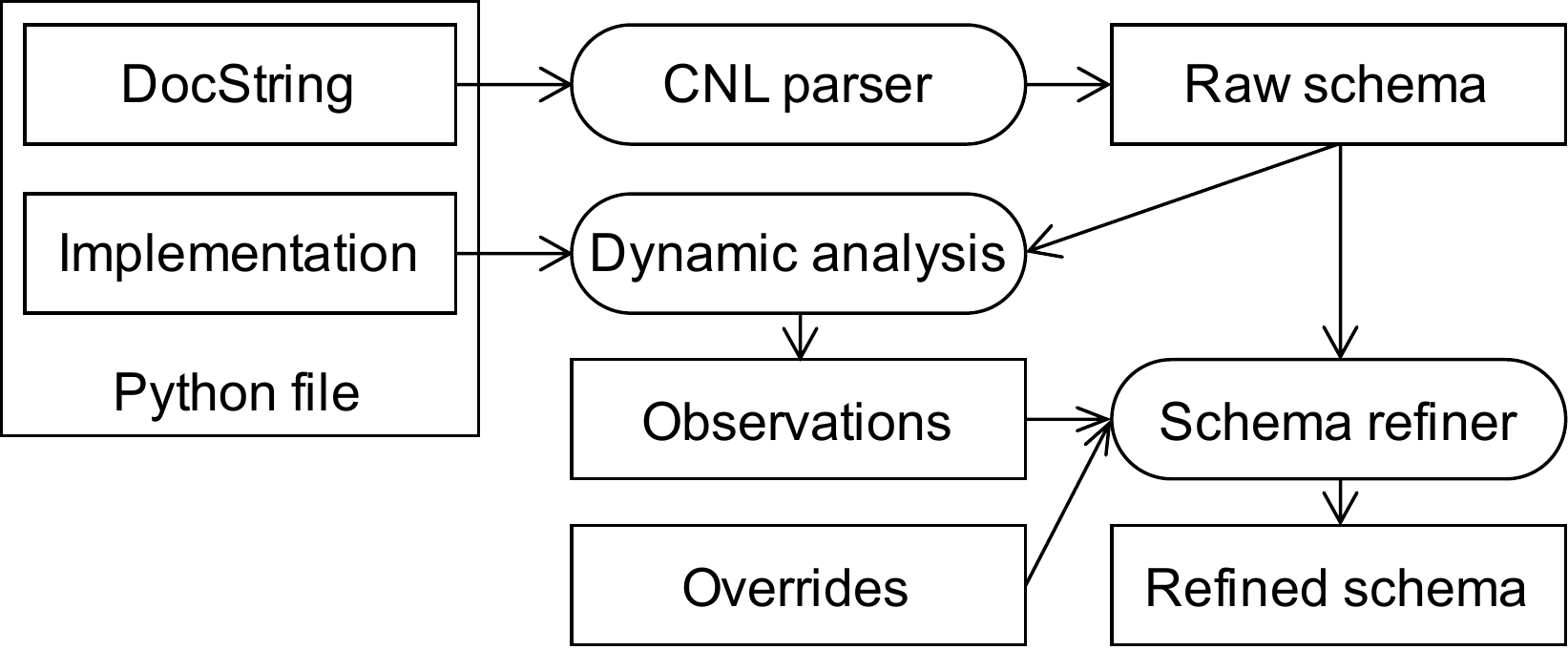}}
  \vspace*{-2mm}
  \caption{\label{fig:overview}Overview of our mining approach.}
\end{wrapfigure}

Python has recently emerged as the dominant ML language and many ML
libraries adopt scikit-learn style conventions for interoperability
(including
PyTorch~\cite{skorch},
pandas~\cite{sklearn-pandas},
Spark~\cite{spark-sklearn},
statsmodels~\cite{statsmodels}, and
TensorFlow~\cite{keras-sklearn}).
This
paper proposes and demonstrates an approach for mining hyperparameter schemas from the
Python file implementing an ML operator. The approach, shown in
Figure~\ref{fig:overview}, mines the docstring and refines the
resulting schema via dynamic analysis of the implementation. Using the
Python file as a single source of truth simplifies maintenance when a
new library version adds new features or deprecates old ones.
Furthermore, since the Python file is written by library developers and the
documentation is widely read by library users, it is a reliable source of
truth.

Our approach outputs hyperparameter schemas in \emph{JSON Schema},
which is a type description language for JSON
documents~\cite{pezoa_et_al_2016}.  JSON Schema is widely adopted for
web APIs, cloud management, and document databases, among other
domains, and there are abundant public resources for learning and
using it.  JSON Schema is independent from specific AI automation
tools and recent work has demonstrated that it can be converted to
specifications for popular such tools~\cite{baudart_et_al_2020}.  We found
JSON Schema to have just the right expressiveness for hyperparameters
including categoricals and conditional constraints.  Furthermore, we
found JSON Schema easy to extend with additional meta-data such as
distributions for continuous hyperparameters.

This paper makes the following contributions:
(1)~Mining Python docstrings to extract hyperparameter schemas including constraints.
(2)~Using dynamic analysis to obtain additional information about hyperparameters beyond the docstrings.
(3)~Reconciling hyperparameter metadata into a single machine-readable schema in JSON Schema format.
We evaluate our approach on \totalOperators automatically mined
hyperparameter schemas for  ML operators and \curatedOperators hand-curated schemas. 
We make both datasets publicly available (\url{https://github.com/IBM/lale/tree/master/lale/lib/}\textsf{\small(autogen$|$sklearn)}).
Overall, we hope this paper contributes towards making HPO tools
easier to use, more reliable, and more effective.

\section{Problem Statement}\label{sec:problem}

This paper is about solving the problem of mining hyperparameter
specifications from a Python docstring and turning them into a JSON
Schema. To make things concrete, \Cref{fig:lr_hp} show an example input and the desired corresponding output of this mining problem.

The left side of \Cref{fig:lr_hp} shows an excerpt of class
\python{sklearn.linear_model.LogisticRegression} with its docstring. A
\emph{docstring} is a string literal that documents a specific class
or function definition. The HTML documentation for scikit-learn and other
popular ML libraries is auto-generated from their docstrings.  For
this to work, the docstrings follow conventions understood by the HTML
generation tool, in this case, the numpydoc
extension~\cite{numpydoc_2008} for Sphinx~\cite{brandl_2008}.  In
other words, docstrings are written in a controlled natural language
(CNL) \cite{kuhn_2014}: controlled, since they follow numpydoc
conventions, and natural, since they are human-readable even before
being converted into HTML. In practice, while docstrings suffice for
HTML generation, they exhibit variability and typos that make schema
extraction non-trivial.

This paper proposes an extractor that converts the docstring not to
HTML but to JSON Schema.  The right side of
\Cref{fig:lr_hp} shows the schema for two categorical arguments
\python{solver} and \python{penalty} and one continuous
arguments \python{C}. 
Like most ML libraries, scikit-learn encodes categorical hyperparameters via Python string constants as opposed to Python enums, but only the values
mentioned in the documentation are valid. 
The example also contains a conditional hyperparameter constraint:
the value of \python{solver} implies which values are valid for
\python{penalty}. We can express this implication by taking the logically
equivalent form
\mbox{\small$\neg\textit{\textsf{premise}}\vee\textit{\textsf{conclusion}}$}
and using the JSON Schema keywords \json{anyOf} and~\json{not}.

\begin{figure}[t]
\begin{minipage}{0.49\textwidth}
\input{lr_code}
\end{minipage}
\begin{minipage}{0.51\textwidth}
\input{lr_schema}
\end{minipage}
\caption{Simplified excerpt of the scikit-learn code for the LogisticRegression estimator~(left) and the corresponding hyperparameters schema (right).}
\label{fig:lr_hp}
\vspace*{-4mm}
\end{figure}



\section{Mining Docstrings}\label{sec:docstrings}

This section describes the CNL Parser component of the overview
diagram in Figure~\ref{fig:overview}.  A CNL (controlled natural
language) is a natural language (e.g., English) with some
amount of structure (e.g., the input format for Sphinx and numpydoc,
adopted by scikit-learn and other ML libraries).
The CNL parser starts by reading the docstring from the Python file,
such as the one on the left of \Cref{fig:lr_hp}. It uses Sphinx
and numpydoc to extract the docstring of methods \python{__init__}
(class constructor), \python{fit} (for training), and
\python{predict} or \python{transform} (for using an operator
after training) of the operator class.
Sphinx and numpydoc pre-parse this information
into a list of argument tuples of the form
\python{(name, short_desc, long_desc)} as well as
descriptions of the return values of the methods. 
Given the list of argument tuples, the
CNL parser uses two hand-crafted CNL grammars to extract per-argument
schemas and inter-argument constraints, respectively.


\paragraph{Mining Argument Schemas.}
The CNL parser uses a grammar (see \Cref{fig:type_grammar_full} in the appendix) to
extract per-argument schemas from the \python{short_desc} fields and inter-argument constraints that appear in the \python{long_desc}.
%
%
%
The example of \Cref{fig:lr_hp} illustrates the main difficulties for mining and
extraction.  The parser needs to ignore noise such as whitespace,
string quotes, or the trailing backslash~(\textbackslash).  
In addition, even within the
same operator, the documentation uses different ways to express the
same thing, e.g., enumerating values in curly braces \lstinline!{...}!
vs.\ using \texttt{or}.
Overall, these difficulties arise from the `N' in
CNL: docstrings use natural language.
Noise is easy to handle using filtering during
lexical analysis. For the other difficulties, our grammar takes
advantage of the `C' in CNL: docstrings use controlled language to the
extent that they follow the conventions encouraged by Sphinx and
numpydoc. 
The grammar thus specifies multiple syntactic alternatives to capture different ways to express the
same thing (e.g., \bnf{enum} or \bnf{default}).

\paragraph{Mining Constraints.}
For inter-argument constraints, the CNL parser first extracts complete
sentences from the long description. Next, it uses regular
expressions to flag possible candidate constraints, for example,
sentences containing the word `only'. Then it parses each candidate
using a grammar that captures common patterns (see
\Cref{fig:constraint_grammar_full} in the appendix).
Unfortunately, there is great variety in how docstrings express conditional
hyperparameter constraints. 
Our grammar is only a first attempt to extract meaningful information.
When the CNL parser fails to parse a sentence flagged as a potential constraint, it puts a
placeholder into the schema with a \json{TODO}
that a human can fill in later.
Having mined both per-argument schemas and inter-argument constraints,
the last step of the CNL parser is to assemble all the pieces into a
single raw schema. 
The resulting JSON Schema is machine-readable and captures the
information in a format suitable both for validation and for search.



\section{Refining Mined Meta-Data}\label{sec:refining}

This section describes the Schema Refiner component of the overview
diagram in Figure~\ref{fig:overview}. This component uses dynamic
analysis on the Python code to make additional observations, which
it combines with heuristics and overrides to turn the raw schema
from the CNL Parser into a refined schema for HPO tools.

\paragraph{Dynamic analysis for default values.}
Non-algorithmic defaults complicate the analysis of types
and values.  
This occurs, for example, when an argument default is appropriated for
purposes other than parameterization.  
To illustrate, it has become
relatively commonplace within scikit-learn to advise users of upcoming
changes in defaults for important arguments by setting the default
value in the constructor's signature to \python{'warn'}
(e.g., Figure~\ref{fig:lr_hp} (left)) to trigger a warning
message.
%
Our dynamic analysis creates an instance by calling the constructor
\python{__init__()} without passing any explicit arguments, then
calls \python{fit} on the resulting instance --- which might assign the argument to its actual algorithmic default value --- and finally
introspects the instance for these (possibly altered) default values
and their types.

\paragraph{Dynamic analysis via runtime exception testing.}
We use the following techniques to harvest good values and filter bad
values for constructor arguments.

\noindent
\emph{Bad values:}
Defaulting all other arguments, we give a deliberately bad value
to the argument under test and capture the exception.  This
exception text usually reports the bad value which is easily
distinguished in the message.  Frequently, the exception text also
reports valid choices for the argument value, using a range of
syntax that we can parse.


\noindent
\emph{Greedy harvesting:}
We allow argument values that are valid for one operator to be tested
for the same argument name in a different operator.  This occasionally
discovers valid values, particularly for under-documented classes.

\noindent
\emph{Sampling:}
Defaulting all other arguments, an argument's range is sampled for
testing for valid values.  If categorical, all values are tested.
Failed values are filtered out for the class.  The message
received for deliberately false values can help to disambiguate
the complaint in the case where it is not known a priori if
the tested value is good or bad.

\noindent
\emph{Bounds testing:} 
With the caveat that some bounds may depend upon
data and the values of other arguments, the bounds of continuous
ranges can be tested individually for validity and
exclusivity, which can be expressed in JSON Schema
via  e.g., \json{'exclusiveMinimum': True}.

\paragraph{Argument Overrides and Relevance to HPO.}
We provide for a dictionary of overrides that allow the automatically
extracted types and ranges to be replaced with user-specified values,
or for the parameter to be excluded from optimization.  
E.g., suppressing the \python{'mae'} choice for \python{'criterion'} on tree regressors because of prolonged fit times, or custom bounds for numeric parameters.
The \python{ForOptimizer} suffix indicates that these are not hard constraints.
This step also sets the \python{distribution} field
(e.g.\ \python{loguniform}) and the \python{relevantToOptimizer}
list, omitting irrelevant arguments such as verbosity.


\section{Results}\label{sec:results}

This sections measures the effectiveness of our extractor tool with three experiments.
(1)~We mined the schema of \totalOperators operators from three different libraries: \sklearnOperators from
scikit-learn~\cite{buitinck_et_al_2013}, 2 from
XGBoost~\cite{chen_guestrin_2016}, and 2 from
LightGBM~\cite{ke_et_al_2017}. 
(2)~We compared the schemas of \curatedOperators operators with manually curated schemas: \sklearnCuratedOperators
from scikit-learn, 2 from XGBoost, and 2 from LightGBM. 
(3)~We used the generated schemas to find three-steps pipelines for \openmlDatasets OpenML datasets with Lale~\cite{baudart_et_al_2020}, an Auto-ML
library that uses hyperparameter search spaces in JSON Schema.

\begin{wraptable}{r}{0.72\textwidth}
\centering
\footnotesize
\caption{\label{table:global_results}Summary of the auto-generated schemas.}
\begin{tabular}{@{}l@{\hspace*{.6em}}r@{\hspace*{.6em}}r@{\hspace*{.6em}}r@{ }l@{\hspace*{.6em}}r@{ }l@{\hspace*{.6em}}r@{ }l@{}}
& total & coverage &\multicolumn{2}{l}{scikit-learn} & \multicolumn{2}{l}{xgboost} & \multicolumn{2}{l}{lightgbm}\\
\toprule
classes& 119& 1.00& 115&& 2&& 2&\\
arguments& 1,867& 1.00& 1,686&& 88&& 93&\\
types& 1,758& 0.94& 1,606 &{\scriptsize(1,490 + 116)}& 77 &{\scriptsize(73 + 4)}& 75 &{\scriptsize(69 + 6)}\\
default& 1,204& 0.64& 1,090 &{\scriptsize(660 + 430)}& 49 &{\scriptsize(13 + 36)}& 65 &{\scriptsize(61 + 4)}\\
range& 399& 0.50& 339 &{\scriptsize(0 + 339)}& 37 &{\scriptsize(0 + 37)}& 23 &{\scriptsize(0 + 23)}\\
constraints& 43& 0.36& 43&{\scriptsize /118}& 0&{\scriptsize /0}& 0&{\scriptsize /2}\\
\bottomrule
\end{tabular}

\end{wraptable}

\paragraph{Complete dataset.}~\\
\Cref{table:global_results} presents the results of the extractor
executed on the complete dataset (see also \Cref{apx:complete}). 
For each category, we report the number mined by the CNL parser and the corrections made by the schema refiner ($\textit{parser} + \textit{refiner}$).
For the constraints we report the number of valid constraints and the number of detected constraints ($\textit{valid}/\textit{detected}$).
Overall, we were able to mine $94\%$ of the~$1,758$ argument types (including the input/output schemas of the \python{fit}, \python{transform}, and \python{predict} methods of all the operators). 
We extracted a default value for $64\%$ of the arguments but default values are not always relevant, e.g., for the input/output type of the \python{fit} or \python{predict} method.
We found a valid range for~$50\%$ of the~$790$ relevant arguments, i.e., numeric arguments (\json{integer} or \json{number}) or \json{string} arguments used to captures \json{enum} values.
Finally, we detected~$120$ constraints but only~$43$ were converted into valid JSON Schema.


\paragraph{Curated dataset.}
\Cref{table:curated_results} presents the results of the comparison (see also \Cref{apx:curated}).
For this experiment, we focused on the arguments to the operator's constructor.
The extractor correctly mined the type for~$81\%$ of the arguments and the default value for~$97\%$.
\begin{wraptable}{r}{0.62\textwidth}
\footnotesize
\centering
\caption{\label{table:curated_results} Auto-generated vs.\ curated schemas.}
\begin{tabular}{@{}l@{\hspace*{.6em}}c@{\hspace*{.6em}}c@{\hspace*{.6em}}c@{\hspace*{.6em}}c@{\hspace*{.6em}}c@{\hspace*{.6em}}c@{}}
& reference & generated & match &precision & recall & $F_1$\\
\toprule


arguments & 452 & 452 & 452 & 1.00 & 1.00 & 1.00\\
types & 452 & 399 & 367 & 0.92 & 0.81 & 0.86\\
defaults & 452 & 441 & 438 & 0.99 & 0.97 & 0.98\\
ranges & 103 & 83 & 83 & 1.00 & 0.81 & 0.89\\
distributions & 166 & 125 & 125 & 1.00 & 0.75 & 0.86\\
constraints & 65 & 20 (/50) & 18 & 0.90 & 0.28 & 0.42\\
\bottomrule
\end{tabular}
\end{wraptable}

The extractor also found a valid range for~$81\%$ of the~$103$ defined ranges in the curated set, and~$75\%$ of the distributions.
Finally, the extractor detected~$50$ of the~$65$ constraints of the curated set. 
Among the detected cons\-traints, $20$ are converted into valid JSON Schema and~$90\%$ of these match the curated schemas.

\paragraph{Auto-ML pipelines.}
To demonstrate the use of our schemas, we use Lale pipelines of the
form \python{preprocessor >> feature_extractor >> classifier}. Then,
we let AutoML automatically select each step from a predefined set of
operators (see details in \Cref{apx:autoai}) and tune its
hyperparameters based on our extracted schemas.
For comparison, we used auto-sklearn~\cite{feurer_et_al_2015} with the
same resource constraints as a baseline: 1h of optimization time and a
timeout of 6mn per trial.
Note that in this comparison, both the framework and the
hyperparameter schemas differ.
The results show that Lale with our auto-generated schemas achieves
similar accuracies as auto-sklearn, a state-of-the-art tool.

\section{Related Work}\label{sec:related}


The most closely related work is jDoctor, which mines javadoc comments
to extract method pre- and post-conditions~\cite{blasi_et_al_2018}.
The results of mining are similar to schemas in that they can capture
argument ranges and even some constraints. But jDoctor focuses on
Java, whereas we focus on Python code without static type annotations
and with string constants. Furthermore, jDoctor focuses on testing,
whereas we focus on AutoML.

Our schema refiner uses dynamic analysis on Python code to augment the
information extracted from docstrings.  Fuzz
testing, also known as \emph{fuzzing}, is a well-established approach
for finding software defects by generating random
inputs~\cite{miller_fredriksen_so_1990}. While our schema refiner is
inspired by fuzzing, its goal is not to find defects but to
extract schemas.

The primary contribution of this paper is the documentation mining, not
the chosen output format. We could have used different formats to
express hyperparameter schemas.
%
Python~3 introduces optional type annotations that can be checked
statically~\cite{vanrossum_lehtosalo_langa_2014}. Unfortunately, since
Python~3 types lack intersection types, string constant types, and
conditional constraints, they are less suitable for HPO.
PCS is a file format for specifying hyperparameter schemas for the
SMAC tool~\cite{hutter_ramage_2015}. PCS is well-suited for HPO and
JSON Schema can be converted to PCS~\cite{baudart_et_al_2020}.


\section{Conclusion}\label{sec:conclusion}

This paper presents a tool that mines Python docstrings of ML
libraries to extract hyperparameter schemas for HPO. The extracted
schemas include names, types, defaults, and descriptions, ranges and
distributions for continuous hyperparameters, enumerations of
constants for categorical hyperparameters, and constraints for
conditional hyperparameters.



\bibliography{bibfile}

\begin{thebibliography}{19}
\providecommand{\natexlab}[1]{#1}
\providecommand{\url}[1]{\texttt{#1}}
\expandafter\ifx\csname urlstyle\endcsname\relax
  \providecommand{\doi}[1]{doi: #1}\else
  \providecommand{\doi}{doi: \begingroup \urlstyle{rm}\Url}\fi

\bibitem[ker()]{keras-sklearn}
tf.keras.wrappers.scikit\_learn.
\newblock URL
  \url{https://www.tensorflow.org/api_docs/python/tf/keras/wrappers/scikit_learn}.
\newblock (Retrieved June, 2020).

\bibitem[skl()]{sklearn-pandas}
{sklearn-pandas}.
\newblock URL \url{https://github.com/scikit-learn-contrib/sklearn-pandas}.
\newblock (Retrieved June, 2020).

\bibitem[sko()]{skorch}
skorch.
\newblock URL \url{https://github.com/skorch-dev/skorch}.
\newblock (Retrieved June, 2020).

\bibitem[spa()]{spark-sklearn}
{spark-skearn}.
\newblock URL \url{https://github.com/databricks/spark-sklearn}.
\newblock (Retrieved June, 2020).

\bibitem[sta()]{statsmodels}
{statsmodels}.
\newblock URL \url{https://github.com/statsmodels/statsmodels}.
\newblock (Retrieved June, 2020).

\bibitem[Baudart et~al.(2020)Baudart, Hirzel, Kate, Ram, and
  Shinnar]{baudart_et_al_2020}
Guillaume Baudart, Martin Hirzel, Kiran Kate, Parikshit Ram, and Avraham
  Shinnar.
\newblock Lale: Consistent automated machine learning.
\newblock In \emph{KDD Workshop on Automation in Machine Learning
  (AutoML@KDD)}, 2020.
\newblock URL \url{https://github.com/ibm/lale}.

\bibitem[Blasi et~al.(2018)Blasi, Goffi, Kuznetsov, Gorla, Ernst, Pezz\`{e},
  and Castellanos]{blasi_et_al_2018}
Arianna Blasi, Alberto Goffi, Konstantin Kuznetsov, Alessandra Gorla,
  Michael~D. Ernst, Mauro Pezz\`{e}, and Sergio~Delgado Castellanos.
\newblock Translating code comments to procedure specifications.
\newblock In \emph{International Symposium on Software Testing and Analysis
  (ISSTA)}, pages 242--253, 2018.
\newblock URL \url{http://doi.acm.org/10.1145/3213846.3213872}.

\bibitem[Brandl(2008)]{brandl_2008}
Georg Brandl.
\newblock Sphinx {Python} documentation generator, 2008.
\newblock URL \url{http://sphinx-doc.org/}.
\newblock (Retrieved June, 2020).

\bibitem[Buitinck et~al.(2013)Buitinck, Louppe, Blondel, Pedregosa, Mueller,
  Grisel, Niculae, Prettenhofer, Gramfort, Grobler, Layton, VanderPlas, Joly,
  Holt, and Varoquaux]{buitinck_et_al_2013}
Lars Buitinck, Gilles Louppe, Mathieu Blondel, Fabian Pedregosa, Andreas
  Mueller, Olivier Grisel, Vlad Niculae, Peter Prettenhofer, Alexandre
  Gramfort, Jaques Grobler, Robert Layton, Jake VanderPlas, Arnaud Joly, Brian
  Holt, and Ga{\"{e}}l Varoquaux.
\newblock {API} design for machine learning software: Experiences from the
  scikit-learn project, 2013.
\newblock URL \url{https://arxiv.org/abs/1309.0238}.

\bibitem[Chen and Guestrin(2016)]{chen_guestrin_2016}
Tianqi Chen and Carlos Guestrin.
\newblock {XGBoost}: A scalable tree boosting system.
\newblock In \emph{Conference on Knowledge Discovery and Data Mining (KDD)},
  pages 785--794, 2016.
\newblock URL \url{http://doi.acm.org/10.1145/2939672.2939785}.

\bibitem[Feurer et~al.(2015)Feurer, Klein, Eggensperger, Springenberg, Blum,
  and Hutter]{feurer_et_al_2015}
Matthias Feurer, Aaron Klein, Katharina Eggensperger, Jost Springenberg, Manuel
  Blum, and Frank Hutter.
\newblock Efficient and robust automated machine learning.
\newblock In \emph{Conference on Neural Information Processing Systems (NIPS)},
  pages 2962--2970, 2015.
\newblock URL
  \url{http://papers.nips.cc/paper/5872-efficient-and-robust-automated-machine-learning}.

\bibitem[Hutter and Ramage(2015)]{hutter_ramage_2015}
Frank Hutter and Steve Ramage.
\newblock Manual for {SMAC} version v2.10.03-master, 2015.
\newblock URL
  \url{https://www.cs.ubc.ca/labs/beta/Projects/SMAC/v2.10.03/manual.pdf}.
\newblock (Retrieved June, 2020).

\bibitem[Ke et~al.(2017)Ke, Meng, Finley, Wang, Chen, Ma, Ye, and
  Liu]{ke_et_al_2017}
Guolin Ke, Qi~Meng, Thomas Finley, Taifeng Wang, Wei Chen, Weidong Ma, Qiwei
  Ye, and Tie-Yan Liu.
\newblock {LightGBM}: A highly efficient gradient boosting decision tree.
\newblock In \emph{Conference on Neural Information Processing Systems (NIPS)},
  pages 3146--3154, 2017.
\newblock URL
  \url{http://papers.nips.cc/paper/6907-lightgbm-a-highly-efficient-gradient-boosting-decision-tree}.

\bibitem[Komer et~al.(2014)Komer, Bergstra, and
  Eliasmith]{komer_bergstra_eliasmith_2014}
Brent Komer, James Bergstra, and Chris Eliasmith.
\newblock Hyperopt-sklearn: Automatic hyperparameter configuration for
  scikit-learn.
\newblock In \emph{Python in Science Conference (SciPy)}, pages 32--37, 2014.
\newblock URL
  \url{http://conference.scipy.org/proceedings/scipy2014/komer.html}.

\bibitem[Kuhn(2014)]{kuhn_2014}
Tobias Kuhn.
\newblock A survey and classification of controlled natural languages.
\newblock \emph{Computational Linguistics}, 40\penalty0 (1):\penalty0 121--170,
  2014.
\newblock URL
  \url{http://www.mitpressjournals.org/doi/pdf/10.1162/COLI_a_00168}.

\bibitem[Miller et~al.(1990)Miller, Fredriksen, and
  So]{miller_fredriksen_so_1990}
Barton~P. Miller, Louis Fredriksen, and Bryan So.
\newblock An empirical study of the reliability of {U}nix utilities.
\newblock \emph{Communications of the ACM (CACM)}, 33\penalty0 (12):\penalty0
  32--44, December 1990.
\newblock URL \url{http://doi.acm.org/10.1145/96267.96279}.

\bibitem[{numpydoc maintainers}(2008)]{numpydoc_2008}
{numpydoc maintainers}.
\newblock numpydoc -- {Numpy}'s {Sphinx} extensions, 2008.
\newblock URL \url{https://numpydoc.readthedocs.io}.
\newblock (Retrieved June, 2020).

\bibitem[Pezoa et~al.(2016)Pezoa, Reutter, Suarez, Ugarte, and
  Vrgo\v{c}]{pezoa_et_al_2016}
Felipe Pezoa, Juan~L. Reutter, Fernando Suarez, Mart\'{\i}n Ugarte, and Domagoj
  Vrgo\v{c}.
\newblock Foundations of {JSON} {Schema}.
\newblock In \emph{International Conference on World Wide Web (WWW)}, pages
  263--273, 2016.
\newblock URL \url{https://doi.org/10.1145/2872427.2883029}.

\bibitem[{van Rossum} et~al.(2014){van Rossum}, Lehtosalo, and
  Langa]{vanrossum_lehtosalo_langa_2014}
Guido {van Rossum}, Jukka Lehtosalo, and Lukasz Langa.
\newblock {PEP} 484 -- type hints, 2014.
\newblock URL \url{https://www.python.org/dev/peps/pep-0484/}.
\newblock (Retrieved June, 2020).

\end{thebibliography}

\clearpage
\appendix

\section{Mining Docstrings}

\subsection{Mining Argument Schemas}
Sphinx and numpydoc pre-parse the documentation
into a list of argument tuples of the form
\python{(name, short_desc, long_desc)}.
The CNL parser uses the grammar in Figure~\ref{fig:type_grammar_full} to
extract per-argument schemas from the \python{short_desc} fields.

The start symbol of the grammar, \bnf{start}, splits the docstring into
three parts:
the type, or a sequence of possible types (encoded in JSON using the \json{anyOf} keyword); an optional flag; and the default value.

\subsection{Mining Constraints}

For inter-argument constraints, the CNL parser first extracts complete sentences from the long description. 
Next, it uses a set of regular expression rules to flag possible candidate constraints, for example, sentences containing the word `only'. 
On each candidate it discovers, the CNL parser uses the grammar in \Cref{fig:constraint_grammar_full}.

Unfortunately, there is great variety in how docstrings express conditional hyperparameter constraints. 
Since our CNL grammar does not anticipate the syntax of this example, it cannot fully extract this constraint.
However, it does detect the presence of some constraint, and it puts a placeholder into the raw JSON Schema with a \json{TODO} that a human domain expert can then fill in later to further enhance it.

\begin{figure}
\begin{lstlisting}[language=bnf]
start    ::= seq optional default ("." | ",")?
seq      ::= (type ","?)+ ("or " type)?

type     ::= "int" | "integer" | "float" | "double" | "boolean" | "bool" | "string" | "str"
           | "None" | "Ignored"  | "callable" | "dict" | "type"
           | obj | array | enum

optional ::= (", optional")?

default  ::= ","? ("default" ("=" | ":")?  val
           | "(" "default" ("=" | ":")? val ")"
           | val "by " "default"
           | "or " val "(" "default" ")")?

obj      ::= "object" | "RandomState" "instance" | "returns an instance of self"

array    ::= atype (shape)?
atype    ::= "list" | "array" | "tuple" | "array_like" | "array-like" 
           | "numpy" "array" | "sparse" "matrix" | "scipy.sparse" |  "scipy" "sparse"
           | "{"? atype ("or " | ",") atype "}"?
shape    ::= ","? "of "? ("shape" | "size")? "="? vtuple ("or " shape)?

enum    ::= "{" val (","? "or "? ("an " | "a ")? val)* "}"
           | ("string" | "str") ","? enum
           | "["? val ("|" val)+ "]"?

vtuple   ::= ("(" | "[") val ("," val)* ","? ("]" | ")") | "None"

val      ::= NAME | NUMBER
\end{lstlisting}
\caption{\label{fig:type_grammar_full}CNL grammar for parsing per-argument schemas.}
\end{figure}
  
\begin{figure}
\begin{lstlisting}[language=bnf]
start    ::= only when cond

only     ::= "only" ("used" | "effective" | "compatible" | "significant" | "available" | "applies")?
when     ::= "when" | "if" | "with" | "in" | "for"

cond     ::= atom | cond ("and" | "or") cond
atom     ::= NAME compare seq | "the"? seq NAME ("is used")?
seq      ::= val (("," val)* ","? ("and" | "or") val)?

compare  ::= "==" | "=" | ">" | "<" | ">=" | ">=" | "is set to" | "is"
val      ::= NUMBER | NAME
\end{lstlisting}
\caption{\label{fig:constraint_grammar_full}CNL grammar for parsing inter-argument constraints.}
\end{figure}

\section{Results}

This section measures the effectiveness of our extractor tool on two
datasets: complete and curated. The \emph{complete dataset} comprises
\totalOperators operators from three different libraries: \sklearnOperators from
scikit-learn~\cite{buitinck_et_al_2013}, 2 from
XGBoost~\cite{chen_guestrin_2016}, and 2 from
LightGBM~\cite{ke_et_al_2017}. 
For XGBoost and LightGBM we considered both the regressor and the classifier.  
For scikit-learn we filter the classes revealed by \python{sklearn.utils.testing.all_estimators()} to obtain estimators and transformers.  
We exclude classes that are abstract, or that are meta-estimators. 
Further we examine their method resolution order, confirm the existence of \python{fit}, and \python{predict} or \python{transform}, and confirm their signatures. 
Finally, we exclude some classes known to be deprecated, e.g., \python{Imputer}, and some known to be intended only to be used by other classes, e.g., \python{ExtraTree}.

The \emph{curated dataset} comprises \curatedOperators operators with manually curated schemas: \sklearnCuratedOperators
from scikit-learn, 2 from XGBoost, and 2 from LightGBM.  Whereas the
complete dataset allows us to evaluate the robustness and coverage of
the tool overall, the curated dataset allows us to compare the
extracted schemas against a ground truth for reference.

Finally, we used the generated schemas to find three-steps pipelines for 15 OpenML datasets with Lale and compare the results with  with auto-sklearn.

\subsection{Complete Dataset Evaluation}\label{apx:complete}

Table~\ref{table:global_results} presents the results of the extractor
executed on the complete dataset.  For each library, it reports the
number of extracted arguments, types, default values, ranges, and constraints.
In addition, Table~\ref{table:global_results} explicitly shows the contributions of the CNL parser and the schema refiner for each category to illustrate the advantages of combining the two strategies.

Overall, we were able to mine $94\%$ of the~$1,867$ argument types (including the input/output schemas of the \python{fit}, \python{transform}, and \python{predict} methods of all the operators).
Missing argument types mostly come from unsupported or under-specified data structures, e.g., \python{object}, \python{dict}, or \python{callable}.
Even when mined, it is not clear how an AI automation tool would
instantiate such arguments during search.

We extracted a default value for $64\%$ of the arguments but default values are not always relevant, e.g., for the input/output type of the \python{fit} or \python{predict} method.
Table~\ref{table:global_results} shows that, even if default values
are often documented, the schema refiner can extract a lot of
additional information, e.g., via dynamic analysis.

Since ranges are not consistently documented, the range analysis is solely based on the schema refiner.
We found a valid range for~$50\%$ of the~$792$ relevant arguments, i.e., numeric arguments (\json{integer} or \json{number}) or \json{string} arguments used to captures \json{enum} values.
However, ranges are not required for all of these arguments.
In fact we observe that, even in the curated schemas, ranges are only defined for a few arguments to produce valid search spaces for hyperparameter search tools.

Finally, we detected~$120$ constraints but only~$43$ were converted into valid JSON Schema. The remaining~$77$ generate \json{TODO} warnings that can be manually inspected.

\subsection{Curated Dataset Evaluation}\label{apx:curated}

Next, we compare the result of the extractor with a set of manually curated schemas. 
Results are presented in Table~\ref{table:curated_results}.
Compared to Table~\ref{table:global_results} we focus on the
arguments to the operator's constructor (\python{__init__}),
leaving aside the arguments and return values of its \python{fit},
\python{predict}, or \python{transform} methods
For each category we report the reference number, the generated
number, and the number of matches between the two sets. We also report
the corresponding precision, recall, and $F_1$ score.

The extractor is able to detect all the~$452$ arguments, which indicates that all the arguments are consistently documented across the three libraries.
In the curated set, all arguments have a correct type and a default value.
The extractor correctly mined the type for~$81\%$ of the arguments and the default value for~$97\%$.
The extractor also found a valid range for~$81\%$ of the~$103$ defined ranges in the curated set.
We categorized a range as valid if the extractor returns an interval
included in the range defined in the curated schema.
Compared to the complete dataset, these counts are relatively low because the curated dataset includes the XGBoost and LightGBM operators that are both more complex and less documented.

Mismatches between extracted and curated default values are due to values that cannot be represented in JSON Schemas: the default value of the \python{missing} argument of the XGBoost operators and the \python{missing_values} of \python{SimpleImputer} is \python{nan}, which is replaced by \python{None} in the curated schemas.
Additionally, our extractor found inconsistencies between the documentation and the code.
For instance, the documentation for the \python{categories} argument of \python{OneHotEncoder} is:
\begin{lstlisting}[language=desc, numbers=none]
categories : 'auto' or a list of lists/arrays of values, \
  default='auto'.
\end{lstlisting}
but the \python{OneHotEncoder.__init__} method gives a default value
\python{categories=None}.

Arguments types are often complex \emph{union} types allowing multiple choices for one argument. For example, \python{n_jobs} can often be either \python{None} or an \json{integer}.
To further investigate the discrepancies between generated and curated types, we analyzed the numbers of \emph{type values} and \emph{enum values}.
The \emph{type values} analysis reports the number of \emph{terminal} types found in the schemas, i.e, \json{boolean}, \json{integer}, \json{number}, \json{string}, or \json{enum}.
The \emph{enum values} analysis reports the number of members found in each \json{enum} list.

\begin{center}
\begin{tabular}{l r r r r r r}
& reference & generated & match &precision & recall & $F_1$\\
\toprule
type values & 631 & 611 & 525 & 0.86 & 0.83 & 0.85\\
enum values & 426 & 269 & 238 & 0.88 & 0.56 & 0.68\\
\bottomrule
\end{tabular}
\end{center}

We observe that for both analyses the precision is relatively high:
$86\%$ for type values and $88\%$ for enum values, which suggests that
extracted data are mostly correct.
However, for enum values the generated number is significantly lower than the curated number: the extractor only found $56\%$ of the enum values in the curated dataset.

This is mostly due to arguments that are under-specified as
\python{string} in the documentation and for which the schema
refiner can not find a suitable enumeration.
For example, the documentation of the \python{criterion} argument
of \python{GradientBoostingClassifier} is: 
\begin{lstlisting}[language=desc, numbers=none]
criterion: string, optional (default="friedman_mse")
\end{lstlisting} 
but the valid enumeration is
\json{['friedman_mse', 'mse', 'mae']}.
These under-specifications are relatively common when the enum value is only one possible choice in a complex union type.
For example, the documentation of the \python{max_features} argument
of \python{DecisionTreeClassifier} is: 
\begin{lstlisting}[language=desc, numbers=none]
max_features: int, float, string or None, optional \
  (default=None)
\end{lstlisting} 
but again the valid enumeration is \json{['auto', 'sqrt', 'log2']}.

Finally, the extractor detected~$50$ of the~$65$ constraints of the curated set. 
Among the detected constraints, $20$ are converted into valid JSON Schema and~$90\%$ of these match the curated schemas.
The mismatches are due to complex constraints that are merged in the curated schemas.
For instance, the description of the \python{power_t} argument of \python{MLPClassifier} contains: 
\begin{lstlisting}[language=desc, numbers=none]
It is used in updating effective learning rate when
the learning_rate is set to 'invscaling'.
Only used when solver='sgd'.
\end{lstlisting}
which is captured in the curated schemas as: "\python{power_t} can only differ from its default value $0.5$ if \python{solver == 'sgd'} and \python{learning_rate == 'invscaling'}", or in JSON Schema:  
\begin{lstlisting}[language=json, numbers=none]
'anyOf': [
    {'type': 'object',
     'properties': {'power_t': {'enum': [0.5]}}}, 
    {'type': 'object',
     'properties': {
        'learning_rate': {'enum': ['invscaling']},
        'solver': {'enum': ['sgd']}}}]}
\end{lstlisting}
but the extractor is only able to extract the second condition on the solver (\python{solver == 'sgd'}).

The results for the constraints show that, even if we are able to flag
most of the constraints, our CNL parser is not the best tool to
extract meaningful information from the constraint candidate.
The language used to express the constraints is far less constrained
than the one used for the type description.  An obvious direction for
future work is thus to try classic natural language understanding
techniques.

\subsection{AutoML Pipelines}\label{apx:autoai}

The selected datasets comprise 5 simple classification tasks (test accuracy~$>90\%$ in all our experiments) and 10 relatively complex tasks (test accuracy~$< 90\%$).
For all the tasks we start from the same three-step pipeline with both generated --- lale-gen in \Cref{tab:accuracy_openml} --- and curated schema --- lale-cur in \Cref{tab:accuracy_openml}.
For comparison, we used auto-sklearn~\cite{feurer_et_al_2015} --- autoskl in \Cref{tab:accuracy_openml} --- as a baseline.
All tasks were configured with the same resource constraints: one hour
of optimization time and a timeout of six minutes per trial.
Lale then uses the search spaces defined in the schemas, the topology of the pipeline, and off-the-self optimizers such as Hyperopt~\cite{komer_bergstra_eliasmith_2014}, to find the best candidate. 

\begin{lstlisting}[language=python]
preprocessors       = [ NoOp, MinMaxScaler, StandardScaler, Normalizer, RobustScaler]
features_extractors = [ NoOp, PCA, PolynomialFeatures, Nystroem]
classifiers         = [ GaussianNB, GradientBoostingClassifier, KNeighborsClassifier,
                        RandomForestClassifier, ExtraTreesClassifier, 
                        QuadraticDiscriminantAnalysis,  PassiveAggressiveClassifier, 
                        DecisionTreeClassifier, LogisticRegression,  XGBClassifier,
                        LGBMClassifier, SVC ]

lale_pipe = make_pipeline( make_choice(*preprocessors), 
                           make_choice(*features_extractors), 
                           make_choice(*classifiers) )               
\end{lstlisting}

For each experiment, we used a $66\%-33\%$ validation-test split, and a 5-fold cross validation on the validation split during optimization.
Experiments were run on a 32 cores (2.0GHz) virtual machine with 128GB memory.
\Cref{tab:accuracy_openml} shows the accuracy results (mean and standard deviation across 5 independent runs) and the number of runs for each experiments (where ``ok'' indicates a successful run and ``ko'' indicates an aborted run).

We observe that Lale with our auto-generated schemas achieves accuracies (88.1\%) that are comparable to auto-sklearn (88.0\% with auto) and Lale with curated schemas (88.5\%).
However, the number of aborted runs show that side-constraints play a key role during the optimization process.
Only 4.0\% of the runs using curated schemas were aborted, compared to 21.7\% with generated schemas (and 6.7\% for auto-sklearn).  

\begin{table}
\begin{center}
  \caption{Accuracy results for the OpenML classification tasks}
  \label{tab:accuracy_openml}
  \begin{small}
  \begin{tabular}{@{}l r@{ }l r@{ }l r@{ }l r@{ }l r@{ }l r@{ }l@{}}
  &	\multicolumn{4}{c}{\textsc{autoskl}} &	\multicolumn{4}{c}{\textsc{lale-gen}} &	\multicolumn{4}{c}{\textsc{lale-cur}} \\
  \cmidrule(lr){2-5}
  \cmidrule(lr){6-9}
  \cmidrule(lr){10-13}
  &	\multicolumn{2}{c}{\textsc{Precision}} &	\multicolumn{2}{c}{\textsc{Runs}} & \multicolumn{2}{c}{\textsc{Precision}} &	\multicolumn{2}{c}{\textsc{Runs}} & \multicolumn{2}{c}{\textsc{Precision}} &	\multicolumn{2}{c}{\textsc{Runs}}\\
  \cmidrule(lr){2-3}
  \cmidrule(lr){4-5}
  \cmidrule(lr){6-7}
  \cmidrule(lr){8-9}
  \cmidrule(lr){10-11}
  \cmidrule(lr){12-13}
  \textsc{dataset}	& \multicolumn{1}{c}{mean} & \multicolumn{1}{c}{std} &	\multicolumn{1}{c}{ok} & \multicolumn{1}{c}{ko} & \multicolumn{1}{c}{mean} & \multicolumn{1}{c}{std} &	\multicolumn{1}{c}{ok} & \multicolumn{1}{c}{ko} & \multicolumn{1}{c}{mean} & \multicolumn{1}{c}{std} &	\multicolumn{1}{c}{ok} & \multicolumn{1}{c}{ko}\\
  \midrule
  \href{https://www.openml.org/d/40981}{australian}	  & 85.09 & (0.39) & 527.8 & (24.8) & 85.35 & (0.45) & 146.0 & (29.2) & 86.14 & (1.02)  & 416.8 & (12.4) \\
  \href{https://www.openml.org/d/1464}{blood}	        & 77.89 & (1.24) & 688.4 & (42.2) & 75.63 & (1.61) & 213.8 & (43.8) & 76.28 & (4.57)  & 351.6 & (9.6)  \\
  \href{https://www.openml.org/d/13}{breast-cancer}	  & 73.05 & (0.52) & 758.6 & (45.2) & 72.63 & (2.40) & 202.4 & (40.4) & 72.00 & (1.43)  & 552.0 & (16.2) \\
  \href{https://www.openml.org/d/40975}{car}	        & 99.37 & (0.09) & 461.6 & (9.0)  & 99.16 & (0.20) & 123.8 & (28.4) & 99.19 & (0.49)  & 138.4 & (9.6)  \\
  \href{https://www.openml.org/d/31}{credit-g}	      & 76.61 & (1.07) & 328.0 & (22.4) & 76.30 & (0.87) & 109.2 & (24.4) & 74.73 & (0.68)  & 223.4 & (11.6) \\
  \href{https://www.openml.org/d/37}{diabetes}	      & 77.01 & (1.18) & 545.0 & (27.4) & 75.83 & (0.81) & 217.8 & (43.8) & 76.77 & (1.61)  & 384.2 & (10.2) \\
  \href{https://www.openml.org/d/1479}{hill-valley}   & 99.45 & (0.87) & 343.0 & (49.6) & 99.65 & (0.30) & 79.4  & (18.2) & 99.60 & (0.20)  & 113.0 & (4.0)  \\
  \href{https://www.openml.org/d/41027}{jungle-chess}	& 88.06 & (0.22) & 28.2  & (8.2)  & 86.66 & (1.06) & 85.8  & (19.4) & 90.29 & (0.00)  & 47.4  & (7.0)  \\
  \href{https://www.openml.org/d/1067}{kc1}	          & 83.79 & (0.28) & 323.4 & (21.4) & 83.19 & (0.27) & 100.4 & (22.0) & 83.28 & (1.10)  & 175.0 & (6.2)  \\
  \href{https://www.openml.org/d/3}{kr-vs-kp}	        & 99.70 & (0.04) & 176.2 & (16.0) & 99.34 & (0.15) & 61.0  & (17.8) & 99.55 & (0.22)  & 79.0  & (8.0)  \\
  \href{https://www.openml.org/d/12}{mfeat-factors}	  & 98.70 & (0.07) & 114.8 & (19.8) & 97.33 & (0.51) & 46.0  & (17.2) & 97.85 & (0.22)  & 62.0  & (12.8) \\
  \href{https://www.openml.org/d/1489}{phoneme}	      & 90.31 & (0.35) & 292.8 & (3.6)  & 88.62 & (0.46) & 129.2 & (24.0) & 88.93 & (0.49)  & 198.4 & (4.6)  \\
  \href{https://www.openml.org/d/40685}{shuttle}	    & 87.27 & (10.3) & 70.8  & (8.6)  & 99.92 & (0.01) & 60.6  & (16.0) & 99.94 & (0.01)  & 49.0  & (8.4)  \\
  \href{https://www.openml.org/d/337}{spectf}	        & 87.93 & (0.77) & 493.4 & (44.4) & 88.10 & (2.58) & 134.0 & (26.2) & 88.10 & (2.14)  & 276.6 & (6.2)  \\
  \href{https://www.openml.org/d/41146}{sylvine}      & 95.42 & (0.19) & 159.8 & (12.2) & 94.46 & (0.19) & 80.0  & (18.2) & 95.14 & (0.14)  & 128.2 & (0.4)  \\
  \end{tabular}
\end{small}
\end{center}
\end{table}

\end{document}